\pdfoutput=1
\documentclass[11pt,a4paper]{article}
\usepackage[hyperref]{acl2021}
\usepackage{multirow}
\usepackage{times}
\usepackage{latexsym}
\usepackage{booktabs}
\usepackage{amssymb}
\usepackage{amsmath}
\usepackage{csquotes}
\usepackage{soul}
\usepackage{graphicx}
\usepackage{tabularx}
\usepackage{float}
\restylefloat{table}
\graphicspath{{figures/}}

\usepackage{microtype}

\aclfinalcopy 
\def\aclpaperid{3120} 

\def\eqref#1{equation~\ref{#1}}

\def\1{\bm{1}}

\DeclareMathAlphabet{\mathsfit}{\encodingdefault}{\sfdefault}{m}{sl}
\SetMathAlphabet{\mathsfit}{bold}{\encodingdefault}{\sfdefault}{bx}{n}

\title{\mntitle}

\author{Machel Reid\thanks{\ \ Corresponding author} \\
  The University of Tokyo \\
  \texttt{\small machelreid@weblab.t.u-tokyo.ac.jp} \\\And
  Victor Zhong \\
  University of Washington \\
  \texttt{vzhong@cs.washington.edu} \\}

\newcommand{\mn}{\tsc{Lewis}}
\newcommand{\mnlong}{\underline{L}evenshtein \underline{e}diting \underline{wi}th unsupervised \underline{s}ynthesis}
\newcommand{\mntitle}{\mn: Levenshtein Editing for Unsupervised Text Style Transfer}
\newcommand{\tsc}[1]{\textsc{#1}}

\newcommand{\MASK}{\texttt{SLOT}\ }

\newcommand{\smallquote}[1]{{\small \texttt{#1}}}

\newcommand{\Yelp}{{\tsc{Yelp}}}
\newcommand{\Amazon}{{\tsc{Amazon}}}
\newcommand{\Polite}{{\tsc{Polite}}}

\newcommand{\uttsrc}{x}
\newcommand{\utttgt}{y}
\newcommand{\editcoarse}{c}

\newcommand{\coarsetagger}{{\rm RoBERTa}_{c}}
\newcommand{\paramcoarse}{{\Phi}_{c}}
\newcommand{\fineeditor}{{\rm BART}_{y}}
\newcommand{\paramfine}{{\Phi}_{{\rm fn}}}

\newcommand{\styletext}{{s}}
\newcommand{\templatetext}{{t}}

\newcommand{\transattn}{{A}}
\newcommand{\poolattn}{{a}}
\newcommand{\poolattnavg}{{\tilde{\poolattn}}}

\newcommand{\synsrc}{{\hat{\uttsrc}}}
\newcommand{\syntgt}{{\hat{\utttgt}}}
\newcommand{\lmsrc}{{\rm BART}_{x}}
\newcommand{\paramsrc}{{\Theta}_{x}}
\newcommand{\lmtgt}{{\rm BART}_{y}}
\newcommand{\paramtgt}{{\Theta}_{y}}

\date{}

\begin{document}

\maketitle
\begin{abstract}
Many types of text style transfer can be achieved with only small, precise edits (e.g.~sentiment transfer from \smallquote{I had a terrible time...} to \smallquote{I had a great time...}).
We propose a coarse-to-fine editor for style transfer that transforms text using Levenshtein edit operations (e.g.~insert, replace, delete).
Unlike prior single-span edit methods, our method concurrently edits multiple spans in the source text.
To train without parallel style text pairs (e.g.~pairs of +/- sentiment statements), we propose an unsupervised data synthesis procedure.
We first convert text to style-agnostic templates using style classifier attention (e.g.~\smallquote{I had a SLOT time...}), then fill in slots in these templates using fine-tuned pretrained language models.
Our method outperforms existing generation and editing style transfer methods on sentiment (\Yelp, \Amazon) and politeness (\Polite) transfer.
In particular, multi-span editing achieves higher performance and more diverse output than single-span editing.
Moreover, compared to previous methods on unsupervised data synthesis, our method results in higher quality parallel style pairs and improves model performance.\footnote{Code and data can be found at \url{https://github.com/machelreid/lewis}}
\end{abstract}

\section{Introduction}

\begin{figure*}[th]
    \centering
    \includegraphics[width=\linewidth]{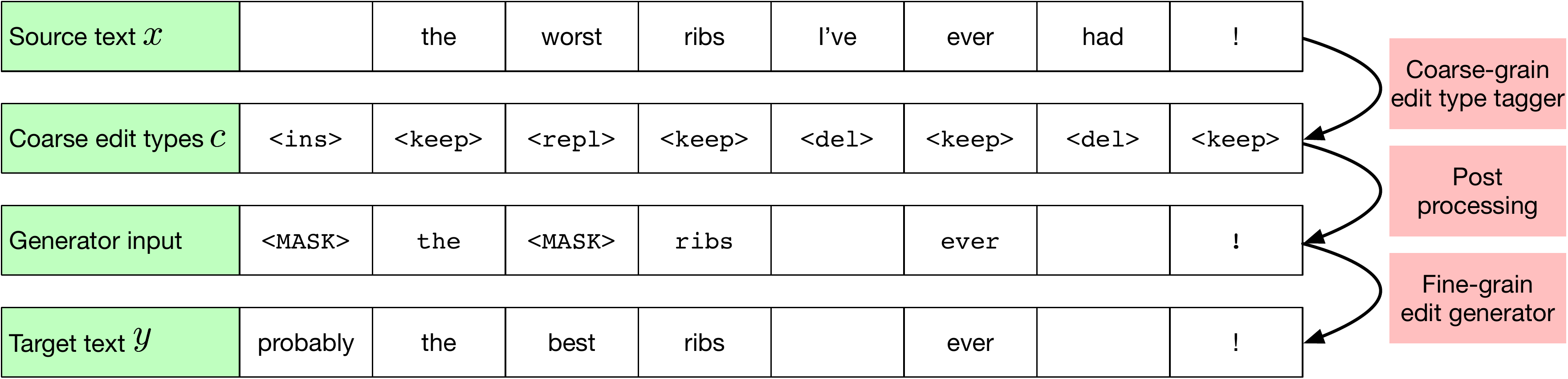}
    \caption{
    Coarse-to-fine Levenshtein editor.
    Given the source text, the two-step editor first generates coarse edit types via a tagger.
    A subsequent generator fills in insertions and replacements while taking into account the source text and the edit types.
    }
    \label{fig:editor}
\end{figure*}

In text style transfer, a model changes the style of a source text (e.g.~sentiment, politeness) into a target style, while otherwise changing as little as possible about the input.
Many types of style transfer can be performed with only small, precise edits instead of generation from scratch.
Consider the task of transforming a negative sentiment sentence such as \smallquote{the worst ribs I've ever had!} to a positive sentence such as \smallquote{probably the best ribs ever!}.
Here, we need only invert the negative sentiment phrase around \smallquote{worst} ---  the references to \smallquote{ribs} should be left as-is.
Recent and concurrent work on text style transfer propose single-span editing~\citep{wu2019mask,malmi-etal-2020-unsupervised}~as an alternative to generating the target text from scratch~\citep{prabhumoye2018style,he2020a,john-etal-2019-disentangled,shen2017tyle,fu2017tyle}.

We introduce a more flexible and powerful multi-span editing method that identifies multiple style-specific components of the text and concurrently edits them into the target style.
Given a source text, we first predict the sequence of coarse-grain Levenshtein edit types (e.g.~insert, replace, delete) that transform the source text to the target text, then fill insertion and replacement edits using a generator.
In the previous example, the operations correspond to inserting the word \smallquote{probably} before \smallquote{the}, replacing \smallquote{worst} with \smallquote{best}, and removing the words \smallquote{I've} and \smallquote{had}.
This example is illustrated in detail in Figure~\ref{fig:editor}.

Learning to edit requires supervised source-target text pairs.
How do we learn high-quality editors when no such supervised parallel data exists?
Given a style text, we synthesize its pair by identifying style-specific content and replacing it with samples from style-specific masked language-models.
In our sentiment transfer example, the style-specific content of the sentence \smallquote{I had a great time at the theatre} is \smallquote{had a great time}.
We can replace this phrase by \smallquote{saw a fantastic movie today} to synthesize an alternative positive-sentiment sentence, or by \smallquote{got ripped off today} to synthesize a negative-sentiment sentence.
Figure~\ref{fig:synthesis} illustrates this example in detail.

We evaluate our editing and synthesis framework, which we call~\mn~(\mnlong), on three style transfer tasks in sentiment (\Yelp, \Amazon) and politeness (\Polite) transfer, and achieve state-of-the-art results in terms of retention of style-agnostic content, similarity to the annotated target text, and transfer accuracy.
\mn~significantly outperforms prior state-of-the-art methods by 2.6-13.5\% accuracy depending on the task.
In further analyses, we show that
(1) compared to concurrent work on editing for style transfer, our editor achieves 33.3\% higher accuracy when trained on the same data;
(2) compared to a competitive BART~\citep{lewis-etal-2020-bart}~pure generation baseline, our editor achieves 5.8\% higher accuracy when trained on the same data;
(3) compared to concurrent work on unsupervised synthesis of style transfer data, our synthesis procedure improves performance by 9.5 BLEU when used to train the same model.
Our experiments show that our editor significantly outperforms both pure generation and editing prior methods, that our editor yields more diverse text transfer and that training on our synthesized data improves performance more than prior synthesis methods.

\section{\mn}

\begin{figure*}[th]
    \centering
    \includegraphics[width=\linewidth]{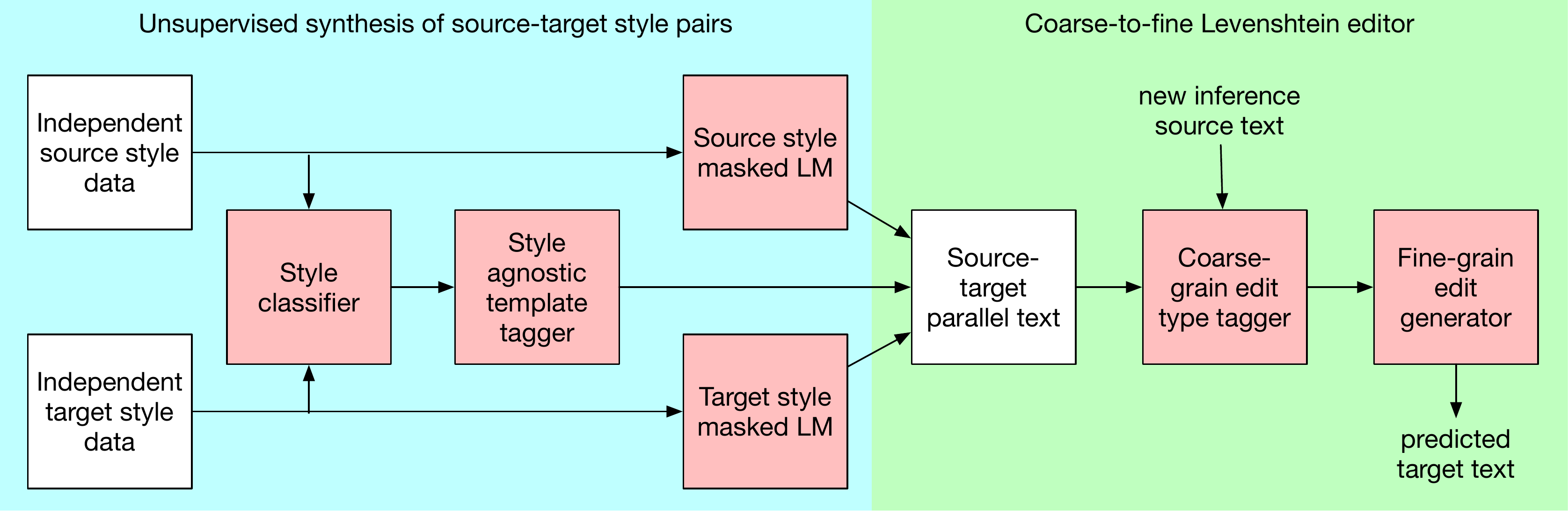}
    \caption{
    \mn~consists of two components.
    Given source-target style text pairs, a coarse-to-fine Levenshtein editor (yellow) first identifies coarse-grain Levenshtein edit types to perform for each token in the source text (e.g.~insert, replace, delete), then fills in the final edits with a fine-grain generator to produce the target text.
    In most applications, supervised source-target style text pairs rarely exist.
    To resolve this lack of annotated data, we perform unsupervised synthesis of source-target style pairs (blue) by first learning to produce style-agnostic templates given arbitrary style text.
    Next, we fill in slots in the template by sampling from style-specific masked language-models.
    In this figure, source and intermediate data are shown in white while model components are shown in red.
    }
    \label{fig:framework}
\end{figure*}

\mn~consists of coarse-to-fine editing and data synthesis.
The editing component, shown in Figure~\ref{fig:editor}, performs local, precise edits of style-specific content of the source text to produce the target text.
The data synthesis component, shown in Figure~\ref{fig:synthesis}, produces supervised source-target text pairs, which do not exist in naturally, to train the editor.
To apply our method to transfer text from a source style to a target style, we first train style-specific masked language models, with which we synthesize source-target text pairs.
We then compute Levenshtein operations for these source-target text pairs and train the coarse-to-fine editor to reproduce these operations.
The full~\mn~is shown in Figure~\ref{fig:framework}.
For ease of exposition, we first describe the editor, then describe how to synthesize parallel data to train the editor.

\subsection{Style transfer via Levenshtein editing}

We propose a coarse-to-fine editor that first predicts coarse-grain Levenshtein edit types~\citep{Levenshtein}, then fills in fine-grain edits with a generator.
Figure~\ref{fig:editor} illustrates the editor.

Suppose we are to transfer text from a source style into a target style.
Let $\uttsrc$ denote the source text, which we would like to edit into the target text $\utttgt$.
In the example shown in Figure~\ref{fig:editor}, we transform the source text \smallquote{the worst ribs I've ever had!} into \smallquote{probably the best ribs ever!}
Our approach has two parts:
The source text is first tagged with a sequence of coarse-grain Levenshtein transition types $\editcoarse$ that transform $\uttsrc$ into $\utttgt$.
A generator then fills in phrases for insertion and replacement operations.
The set of coarse Levenshtein transition types are \smallquote{insert}, \smallquote{keep}, \smallquote{replace}, and \smallquote{delete}.
In the running example, the sequence of operations are to insert before \smallquote{the}, replace \smallquote{worst}, and delete \smallquote{I've} and \smallquote{had}.

First, we train a RoBERTa-tagger~\citep{liu2019oberta} to generate these coarse edit types, which produces coarse edit types for each token in the source text.
To accommodate the insertion operation, we produce two tags for each token.
The first tag is a binary indicator of whether an additional phrase should be inserted before this token.
The second tag is the non-insertion operation to take for this token.
In the previous example, for instance, the word \smallquote{the} triggers both insertion and keep operations.
\begin{align}
    \editcoarse &= \coarsetagger(\uttsrc; \paramcoarse)
\end{align}
Next, we train a fine-grain edit generator to produce the target text.
Unlike the coarse-grain edit type generator which only observes the source text, the fine-grain edit generator observes both the original source text and the source text with the coarse-grain edit types applied $\uttsrc_\editcoarse$.
We use the edit types produced by the Levenshtein algorithm during training and the edit types predicted by the RoBERTa-tagger during inference.
\begin{align}
    \utttgt &= \fineeditor(\uttsrc, \uttsrc_\editcoarse; \paramfine)
\end{align}
Our generator is a BART-based~\citep{lewis-etal-2020-bart} masked sequence-to-sequence model.
The input to BART is the concatenation of the original source text $\uttsrc$ and the source text with the coarse-grain edit types applied $\uttsrc_\editcoarse$.
The generator is trained to fill in phrases for coarse-grain edit types \smallquote{<ins>} and \smallquote{<repl>}.
In the example,~$\fineeditor$ is given the input text \smallquote{the worst ribs I've ever had! SEP <MASK> the <MASK> ribs ever !} and respectively fills in the two \smallquote{<MASK>}s with \smallquote{probably} and \smallquote{best}.

\subsection{Unsupervised synthesis of source-target style pairs} \label{sec:method.synthesis}

\begin{figure*}[th]
    \centering
    \includegraphics[width=\linewidth]{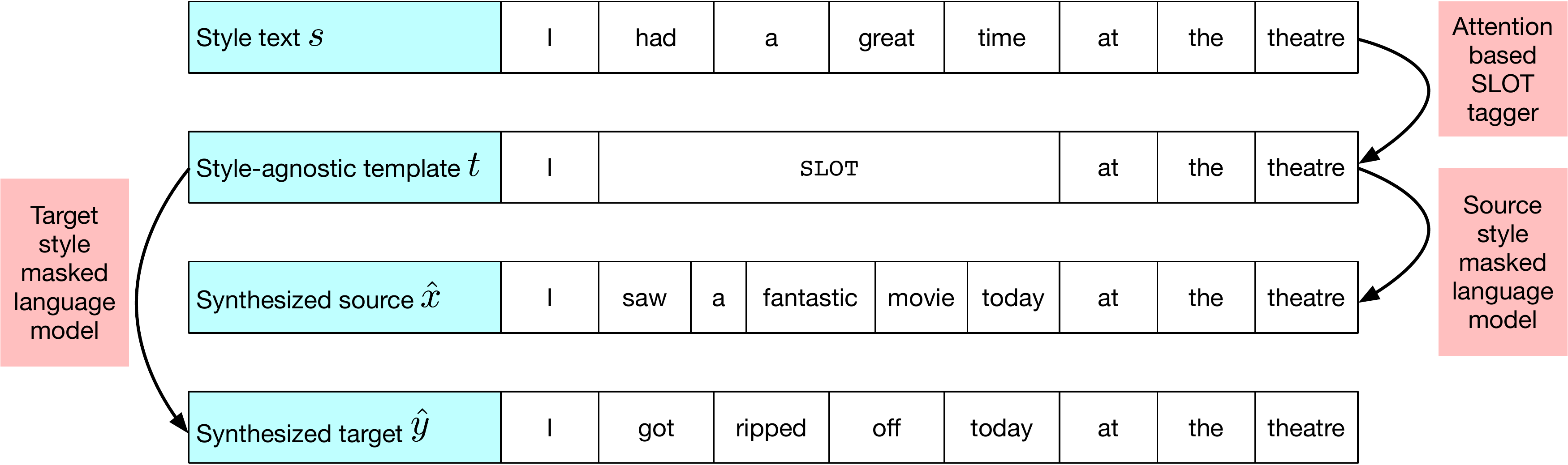}
    \caption{
    Unsupervised synthesis of source-target style pairs.
    We first train an attentive style classifier, whose attention weights we use to identify style-specific content.
    Next, we remove style-specific content with slots to form a style-agnostic template.
    This template is finally filled using style-specific masked language-models for each style to synthesize parallel style text pairs.
    }
    \label{fig:synthesis}
\end{figure*}

Training an editor requires large quantities of source-target text pairs.
While there exists an abundant amount of style-specific data, parallel source-target pairs are difficult to collect and annotate.
How do we train editing style transfer methods when no such data exists?
We hypothesize that pretrained masked language-models, when carefully constrained to generate only style-specific content, can provide high-quality source-target pairs for style transfer.

Our synthesis procedure, shown in Figure~\ref{fig:synthesis}, is two-fold.
First, given a text $\styletext$ from either style, we identify a style-agnostic template $\templatetext$, in which style-specific content are replaced with slots.
For instance, for the style text \smallquote{I had a great time at the theatre}, the style-agnostic template is \smallquote{I SLOT at the theatre}.
To identify style-specific content, we train a RoBERTa-based style classifier that differentiates between text from each style.
\citet{vaswani2017ttention}~and \citet{hoover-etal-2020-exbert}~show that heavily attended-tokens correlate strongly with tokens that are indicative of the target class.
We observe similar results when inspecting the attention matrices computed by the 12-layer Transfomer for the sentiment classification task.
Namely, the penultimate layer's attention weights correlate strongly with words humans identify as strongly indicative of positive vs. negative sentiment.
Hence, we define style-specific content as tokens that have higher-than-average attention weights in the classifier.

Consider the multi-head attention matrix $\transattn$ in the penultimate Transformer layer, where $\transattn_{ij}$ represents the attention weight of the $j$th attention head on the $i$'s token, normalized across all tokens.
First, we max-pool $\transattn_i$ over all attention heads to form $\poolattn_i$.
Conceptually, $\poolattn_i$ represents the maximum extent to which the $i$th word was attended to by any attention head.
\begin{align}
\poolattn_i = \max_j \transattn_{ij}
\end{align}
Let $N$ denote the sequence length.
We compute the average attention weight $\poolattnavg$ as
\begin{align}
\poolattnavg &= \frac{\sum^N_{i} \poolattn_i}{N}
\end{align}
To modify the style text $\styletext$ into the style-agnostic template $\templatetext$, we keep tokens that have above-average attention weight.
\begin{align}
    \templatetext_{i} &= \begin{cases} \label{eq:attn}
    \MASK, & \text{if}\ \poolattn_i \geq \poolattnavg\\ 
    \styletext_i, & \text{if}\ \poolattn_i < \poolattnavg\ \end{cases}
\end{align}
We merge consecutive $\MASK$ tokens in $\templatetext$.
In the running example, for the style text \smallquote{I had a great time at the theatre}, the tagger generates \smallquote{I SLOT SLOT SLOT SLOT at the theatre}, which after merging becomes \smallquote{I SLOT at the theatre}.

We then fine-tune style-specific masked language-models $\lmsrc$ and $\lmtgt$ to fill in slots in the template and recover the style-specific text.
During training, phrases in the input sentence are randomly discarded and the model is trained to fill the phrases back in~\citep{lewis-etal-2020-bart}.
Having trained style-specific masked language-models for both the source and target styles, we use both models to generate source and target filled-in text given style-agnostic templates.
\begin{align}
    \synsrc &= \lmsrc(\templatetext; \paramsrc) \\
    \syntgt &= \lmtgt(\templatetext; \paramtgt)
\end{align}
In our running example, sampling with the positive language model yields the sentence \smallquote{I saw a fantastic movie today at the theatre}, while sampling with the negative language model yields the sentence \smallquote{I got ripped off today at the theatre}.

The last step we perform is a filtering step using the classifier.
For synthesized examples in style $k$, we keep examples for which the style classifier predicts $k$.
In other words, we keep only examples where the language models and the classifier agree.
We find that this improves data quality and editor performance.
We use the collection of synthesized source and target text pairs $\synsrc$, $\syntgt$ to train the editor.

\section{Experimental Setup}

We focus on two types of text style transfer: (1) \textbf{Sentiment transfer}, in which we transform a positive sentiment sentence to a corresponding negative sentiment sentence or vice-versa without changing the core content (i.e.~attributes of the sentence not concerned with sentiment) (2) \textbf{Politeness transfer}, in which we transform the tone of a sentence from impolite to polite.

\begin{table}[th]
    \centering
    \footnotesize
    \begin{tabularx}{\linewidth}{lXXXX}
    \toprule[0.13em]
         \textbf{\tsc{Dataset}} & \textbf{Attributes} & \textbf{Train} & \textbf{Valid} & \textbf{Test}\\
         \midrule[0.07em]
         \multirow{2}{*}{\Yelp}  & Positive & 270K & 2000 & 500 \\
        & Negative& 180K & 2000 & 500 \\ 
        \midrule[0.05em]
         \multirow{2}{*}{\Amazon} & Positive & 277K & 985 & 500 \\
         & Negative & 278K & 1015 & 500 \\
         \midrule[0.05em]
         \multirow{2}{*}{\Polite} & Polite & 219K &  26K & 800 \\
         & Impolite & 198K &  24K & ---\\
         \bottomrule[0.13em]
    \end{tabularx}
    \caption{Dataset statistics for style transfer tasks. The politeness corpus does not have parallel evaluation data and only evaluates on transfer from impolite to polite.}
    \label{tab:datasets}
\end{table}
\begin{table*}[th]
    \centering
    \begin{tabularx}{\linewidth}{lXXXXX}
    \toprule[0.15em]
         \textbf{Model} &  \textbf{Acc} & \textbf{SBLEU} & \textbf{BLEU} & \textbf{SBERT} & \textbf{BERT} \\
         \midrule[0.07em]
         \textbf{Baselines} &&&&&\\
         Input Copy & 1.5 &  100.0 & 24.8 & 100.0 & 53.74 \\
         Reference & 81.6 &25.3&100.0&53.7& 100.0\\
         \midrule[0.07em] 
         \textbf{Generation methods} &&&&&\\
         Delete and Retrieve~\citep{li-etal-2018-delete} &88.6&36.8&12.2&48.5& 33.3\\
         Tag and Generate~\citep{madaan-etal-2020-politeness} & 86.2 & 47.1 & 19.8 & 57.9 & 37.2 \\
         DeepLatentSeq~\citep{he2020a} & 83.8 & 48.4 & 18.7 & 57.9 & 36.0 \\
         \midrule[0.07em]
         \textbf{Editing methods} &&&&&\\
         Masker~\citep{malmi-etal-2020-unsupervised} & 40.9$^\dagger$ & --- & 14.5 & --- & ---\\ 
         LaserTagger~\citep{malmi-etal-2019-encode} + Masker data &  49.6$^\dagger$ & --- & 15.3 & --- &--- \\
         \midrule[0.1em]
         LaserTagger + our data & 59.8 & \textbf{71.8} & \textbf{24.8} & \textbf{81.3} & \textbf{51.6} \\
         \mn & \textbf{93.1} & 58.5 & 24.0 & 72.2 & 50.0 \\
         \bottomrule[0.15em]
    \end{tabularx}
    \caption{Results on \Yelp. Results with $^\dagger$ are taken from the classifier trained in \citet{malmi-etal-2020-unsupervised} because the outputs for these models are not released.}
    \label{tab:yelp}
\end{table*}

\begin{table*}[th]
    \centering
    \begin{tabularx}{\linewidth}{lXXXXX}
    \toprule[0.15em]
         \textbf{Model} &  \textbf{Accuracy} & \textbf{SBLEU} & \textbf{BLEU} & \textbf{SBERT} & \textbf{BERT} \\
         \midrule[0.07em]
         Delete and Retrieve \citep{li-etal-2018-delete} &51.2&57.1&29.9&66.9& 46.2\\
         Tag and Generate \citep{madaan-etal-2020-politeness} & 60.8 & \textbf{68.7} & \textbf{34.8} & 69.5 & 48.2 \\
         \midrule[0.07em]
         \mn 
         & \textbf{74.3} & 65.6 & 32.9 & \textbf{75.2} & \textbf{52.2} \\
         \bottomrule[0.15em]
    \end{tabularx}
    \caption{Results on \Amazon}
    \label{tab:amazon}
    \vspace{-0.05in}
\end{table*}

%

We make use of three datasets: \Yelp~ \citep{shen2017tyle} consists of 450K sentences from business reviews for training and 1000 sentences released by \citet{li-etal-2018-delete} for testing, \Amazon~\citep{he2016ps} consists of 540K sentences from product reviews for training and 1000 sentences for testing, and \Polite~ \citep{madaan-etal-2020-politeness}, produced by filtering through the Enron Email corpus, consisting of 420K sentences for training and 800 sentences for testing.
We list dataset statistics in Table~\ref{tab:datasets}.

\subsection{Training Setup}
We implement our models using \texttt{fairseq}\footnote{\url{https://github.com/pytorch/fairseq}} \citep{ott-etal-2019-fairseq} and HuggingFace\footnote{\url{https://github.com/huggingface/transformers}} \citep{wolf-etal-2020-transformers} --- both based on the PyTorch library \citep{pytorch}.
For BART-based generation models, we initialize with BART-base \citep{lewis-etal-2020-bart}, and train using a batch size of 65K tokens for 30000 iterations. We use a linear warmup schedule, reaching the peak of $3 \times 10^{-5}$ at 5000 iterations, and then proceed to decay the learning rate with a polynomial decay schedule. For regularization, we use a dropout value of $0.3$ and a weight decay value of $0.1$. We optimize using Adam, with hyperparameters $\beta_1 = 0.9, \beta_2=0.98$ and cross entropy loss.
For RoBERTa-based taggers and classifiers, we initialize with RoBERTa-base~\citep{liu2019oberta}, and train using a batch size of 256 for 5000 iterations. We optimize using Adam, warm up the learning rate to $1\times10^{-6}$ and then decay with a cosine schedule.
We train all models using mixed precision \citep{micikevicius2018mixed} for faster training.
Similar to prior work~\citep{wu2019mask,malmi-etal-2020-unsupervised}, we decode using a beam width of 5 and rerank outputs produced by beam search using the likelihood of the classifier trained in Section~\ref{sec:method.synthesis}.
\begin{table*}[th]
    \centering
    \begin{tabularx}{\linewidth}{lXXX}
    \toprule[0.15em]
         \textbf{Model} &  \textbf{Accuracy} & \textbf{SBLEU} & \textbf{SBERT} \\
         \midrule[0.07em]
         Delete, Retrieve, Generate \citep{li-etal-2018-delete} & --- & 11.8 &  ---\\
         Tag and Generate \citep{madaan-etal-2020-politeness} & 84.8 & 70.4 &  71.6 \\
         \midrule[0.07em]
         \mn &  \textbf{87.4} & \textbf{75.3} & \textbf{81.4} \\
         \bottomrule[0.15em]
    \end{tabularx}
    \caption{Results on \Polite} 
    \label{tab:politeness}
\end{table*}
\begin{table*}[th]
    \centering
    \begin{tabularx}{\linewidth}{lXXXXX}
        \toprule
             \textbf{Model} & \textbf{Acc} &\textbf{SBLEU}& \textbf{BLEU} & \textbf{SBERT}&\textbf{BERT} \\
         \midrule
         LM fill                & 90.3 & 42.9 & 17.4 & 58.9 & 41.6 \\
         Seq2Seq                & 87.3 & 50.0 & 19.3 & 68.5 & \textbf{50.0} \\
         \mn~w/o filtering      & 91.2 & 50.1 & 23.3 & 69.8 & 48.0 \\
         \mn                    & \textbf{93.1} & \textbf{58.5} & \textbf{24.0} & \textbf{72.2} & \textbf{50.0} \\
         \bottomrule
    \end{tabularx}
    \caption{
    Ablation on \Yelp.
    ``LM fill'' is the ablation experiment in which we convert the source style text to a style-agnostic template and directly use the target style language model to synthesize a target style text (e.g.~the editor is not used).
    ``Seq2Seq'' is a pretrained BART model that is fine-tuned on the synthesized data (e.g.~a from-scratch generation model trained on the same data as the editor).
    }
    \label{tab:ablation}
    \vspace{-0.05in}
\end{table*}

\subsection{Comparison with existing methods}
We compare~\mn~to five prior methods:
Delete, Retrieve, Generate \citep{li-etal-2018-delete}, a retrieval method that finds text from the target domain corpus whose style-agnostic form is similar to that of the source text;
Tag and Generate \citep{madaan-etal-2020-politeness}, a generation method that conditionally generates target text from style-agnostic source text;
DeepLatentSeq \citep{he2020a}, an unsupervised machine translation-based approach where generators in each domain are regularized by a language model-based latent prior.
Finally, we compare to previous editing approaches proposed by \citet{malmi-etal-2019-encode, malmi-etal-2020-unsupervised} where a single span in the source text two domain-specific language models disagree on is replaced.

\subsection{Evaluation}
\paragraph{Automatic Evaluation} We use five evaluation metrics:
BLEU \citep{papineni-etal-2002-bleu} measured against the reference (denoted as \textbf{BLEU}) to evaluate lexical overlap with human annotation;
Self-BLEU measured against the source to measure content preservation (denoted as \textbf{SBLEU});
BERTScore and Self-BERTScore \citep{zhang2020bert} measured against the reference and the source (denoted as \textbf{BERT} and \textbf{SBERT} respectively);
and accuracy measured against an external classifier (denoted as \textbf{Accuracy}) to measure how well the style was transferred.
\begin{table}[H]
    \centering
    \footnotesize
    \begin{tabularx}{\linewidth}{llXX}
         \toprule[0.1em]
         && \textbf{BLEU} & \textbf{BERT} \\
         \midrule[0.05em] 
         \textbf{Ref} & \multicolumn{1}{l}{great place , great food !} & --- & --- \\
         \textbf{Hyp 1} & pathetic place , great food ! & \textbf{76.0} & 71.5\\
         \textbf{Hyp 2} &  amazing place , awesome food ! & 0.0 & \textbf{78.6}\\ 
         \bottomrule[0.1em]
    \end{tabularx}
    \caption{Example comparing BERTScore vs BLEU. \textbf{Ref} denotes the reference sentence and \textbf{Hyp 1} and \textbf{Hyp 2} represent two example hypotheses.}
    \label{tb:bertscore_demo}
\end{table}

While measuring BLEU, Self-BLEU, and accuracy are standard for this task, we propose additionally using BERTScore due to its higher correlation with human judgments \citep{zhang2020bert}. Compared to BLEU and Self-BLEU which are n-gram based, BERTScore is measured using token-wise cosine similarity between representations produced by BERT \citep{devlin-etal-2019-bert}. 

Given this, the usage of BERTScore addresses the potential issue of accurately transferred sentences being scored poorly due to its low n-gram overlap. Table~\ref{tb:bertscore_demo} shows an example of this where the style is accurately transferred but is scored poorly by BLEU as a result of low n-gram overlap.

Furthermore, following \citet{malmi-etal-2020-unsupervised} who use a BERT-based classifier to score their outputs, we train a classifier initialized with RoBERTa-base \citep{liu2019oberta}. 
This model correctly classifies 98.2\% of the \Yelp~classification test set by~\citet{shen2017tyle}.
Its accuracy is used to evaluate the output of style transfer models.

\paragraph{Human Evaluation} We perform a robust human evaluation on all datasets, asking crowdworkers to rate 300 examples from Yelp (150 positive, 150 negative), 200 examples from Amazon (100 positive,100 negative) and 100 from Politeness. Five annotators rate each pair from 1 (strongly disagree) to 5(strongly agree) in terms of fluency, content preservation (CP) and style transfer. We compare with our strongest baseline Tag and Generate \citep{madaan-etal-2020-politeness}.

\begin{table}[]
    \centering
    \resizebox{0.5\textwidth}{!}{
    \begin{tabular}{lcccc}
    \toprule
         \textbf{Dataset} &\textbf{Model} & \textbf{Fluency} & \textbf{CP} & \textbf{Style} \\
         \midrule
         \multirow{2}{*}{\Yelp} & TG & 3.84$\pm$1.01 & 3.63$\pm$0.93 & 3.67$\pm$1.02\\
         & \mn & \textbf{3.94}$\pm$0.99 & \textbf{3.76}$\pm$0.88 & \textbf{3.72}$\pm$0.98  \\
         \midrule
         \multirow{2}{*}{\Amazon} & TG & 3.60$\pm$1.01&3.48$\pm$0.93&3.37$\pm$1.02\\
         & \mn & \textbf{3.65}$\pm$0.88 & \textbf{3.50}$\pm$0.88 & 3.37$\pm$0.90 \\
         \midrule
         \multirow{2}{*}{\Polite} & TG & 3.83$\pm$0.84&3.76$\pm$0.90&3.48$\pm$1.04\\
         & \mn &\textbf{3.93}$\pm$0.78&\textbf{3.87}$\pm$0.83&\textbf{3.63}$\pm$0.98 \\
         \bottomrule
    \end{tabular}}
    \caption{Human evaluation results comparing \mn~and Tag and Generate (TG)}
    \label{tab:my_label}
\end{table}

\section{Results}
\begin{figure*}[th]
    \centering
    \includegraphics[width=\linewidth]{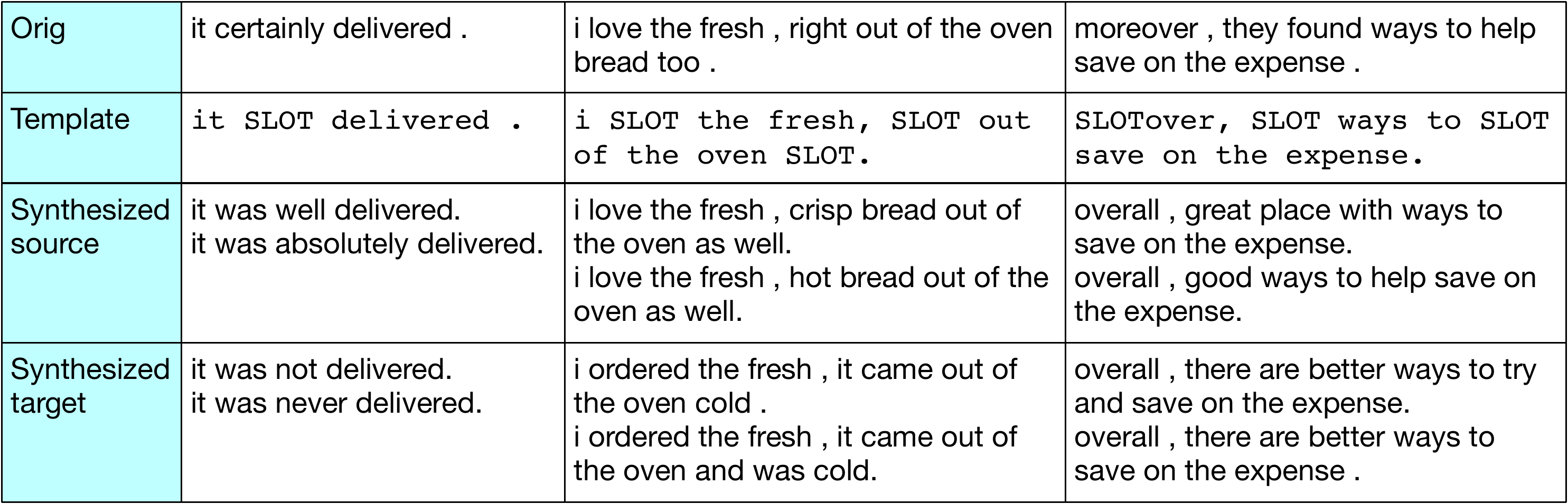}
    \caption{Examples of synthesized parallel text on the \Yelp~dataset.}
    \label{fig:ex_synthesis}
\end{figure*}

Performance of~\mn~compared to other methods on \Yelp, \Amazon, and \Polite~are respectively shown in Tables~\ref{tab:yelp},~\ref{tab:amazon}, and~\ref{tab:politeness}, with human evaluation shown in Table~\ref{tab:my_label}.
\mn~outperforms prior methods on all datasets in terms of accuracy, BLEU, and BERTScore:
\mn~achieves more successful transfers (2.6-13.5\% accuracy depending on task), has higher overlap with human annotations (4-14.4 BERTScore), and retains more source content (5.7-14.3 Self-BERTScore). Human evaluation ($p=0.01$ for Yelp/Polite using pairwise bootstrap sampling \citep{koehn-2004-statistical}) shows that \mn~outperforms Tag and Generate on fluency, content preservation and style across datasets.
These results indicate that~\mn~is an effective method for style transfer.
On the \Amazon~dataset --- which is noisier than the \Yelp~dataset --- \mn~underperforms Tag and Generate when evaluating using BLEU, however when evaluating using BERTScore~\mn~outperforms the latter.
When we inspect the output of~\mn, we find that it generates more diverse output as shown in Figure~\ref{fig:ex_synthesis}.

\begin{table}[t]
\centering
\begin{tabular}{@{}lll@{}}
\toprule
                                & Mean  & Std  \\ \midrule
\# merged edit ops              & 1.57  & 0.78 \\
\# source toks                  & 10.74 & 2.73 \\
\# style-agnostic template toks & 10.42 & 2.86 \\
\# edit output toks             & 11.24 & 3.16 \\ \bottomrule
\end{tabular}
\caption{Coarse-to-fine editor statistics on \Yelp, after merging consecutive edit operations of the same type, so that the number of operations denote spans as opposed to tokens (e.g. delete, replace).}
\label{tab:edit_stats}
\vspace{-0.05in}
\end{table}

\begin{figure*}[th]
    \centering
    \includegraphics[width=\linewidth]{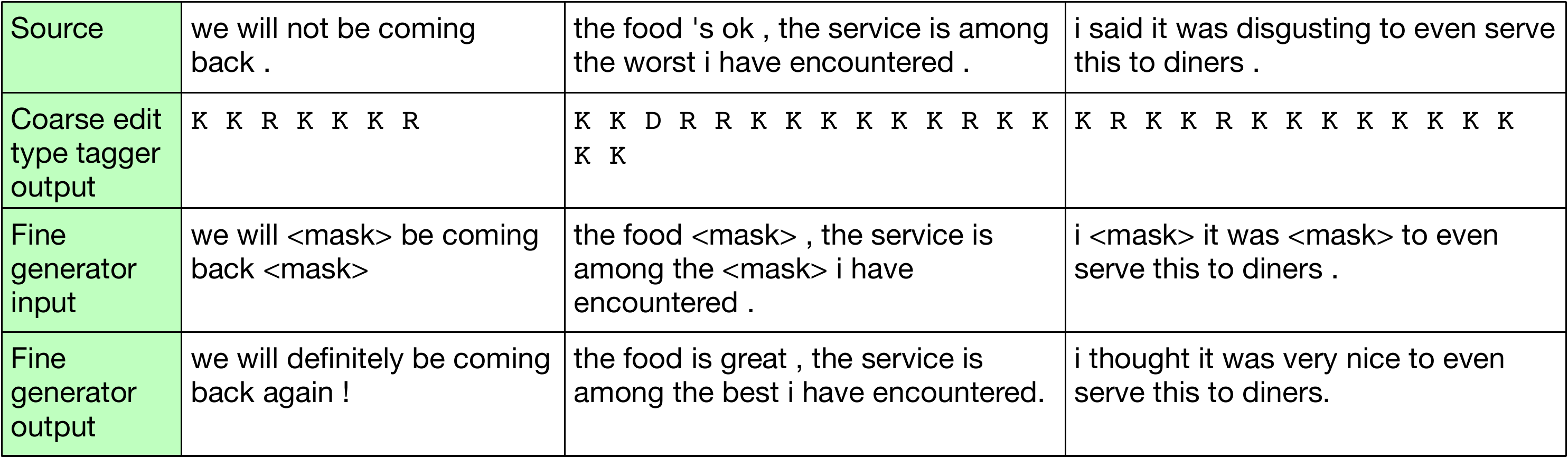}
    \caption{Examples of coarse-to-fine editor output on the \Yelp~dataset. We abbreviate the edit operation with K for \smallquote{<keep>}, D for \smallquote{<del>}, and R for \smallquote{<repl>}. Unlike previous and concurrent edit methods, we concurrently edit multiple spans in the text.}
    \label{fig:ex_edit}
    \vspace{-0.2in}
\end{figure*}

One reason that~\mn~generates more diverse output is that unlike previous and concurrent editing work that use single-span replacement~\citep{malmi-etal-2019-encode,malmi-etal-2020-unsupervised}, our method concurrently edits multiple spans with a larger set of operations.
This is inherently supported by the editor (Figure~\ref{fig:ex_edit}) as well as encouraged during unsupervised data synthesis (Figure~\ref{fig:ex_synthesis}).
Table~\ref{tab:edit_stats}~shows that a large number of examples do require multiple edits, and that the coarse-to-fine editor indeed performs multiple edit operations on average.

In addition to comparing end-to-end systems, we also compare~\mn~to concurrent editing and synthesis methods by~\citet{malmi-etal-2019-encode,malmi-etal-2020-unsupervised}.
Table~\ref{tab:yelp} shows that training the same model (LaserTagger) on our data improves and BLEU by 9.5 (the accuracy difference is not directly comparable since~\citet{malmi-etal-2020-unsupervised}~used a BERT classifier and did not release model output).
This suggests that our data synthesis procedure produces higher quality data than~\citet{malmi-etal-2020-unsupervised}.
Furthermore, because LaserTagger only performs single-span edits, it often fails to transfer the style of the text.
This also accounts for its high BLEU and BERTScore but low accuracy, as we show that a model that simply copies the input also achieves high BLEU and BERTScore but low accuracy.
Replacing LaserTagger with our coarse-to-fine Levenshtein editor results in a sizable 33\% gain in accuracy.
In Table 1 of the Appendix, we show example outputs of these models for comparison.
Finally, we ablate~\mn~to investigate how the different components of~\mn~affect performance.


%
\paragraph{Editing outperforms pure generation}
We replace the coarse-to-fine editor with a sequence-to-sequence BART model, which we also train with synthesized data.
This is a strong baseline that outperforms prior pure generation work on style transfer, as shown in Table~\ref{tab:amazon}.
Nevertheless, Table~\ref{tab:ablation} shows that~\mn~outperforms this baseline on all metrics.
This confirms our hypothesis that editing is a more effective means of style transfer compared to pure generation.

\paragraph{Training on synthesized data improves performance.}
Instead of training an editor using synthesized data, given a source text during inference, we convert it to a style-agnostic template and immediately fill it using the target language model.
Table~\ref{tab:ablation}~shows that the resulting model underperforms both the sequence-to-sequence BART and the coarse-to-fine editor on all metrics.
This result may be surprising, in that one expects the performance of a model trained on data synthesized by language models to be at-most on par with the performance of the language model.
In this case, we observe that training on the synthesized data actually improves over just using the language models.
We hypothesize that this gain is due to the editor learning correlations between the source language model and the target language model, namely how to precisely transform the output of the source language model to the output of the target language model.
The gains we observe here may be related to gains from training on back-translated or pseudo-parallel data~\citep{sennrich-etal-2016-improving,edunov-etal-2018-understanding,he2020revisiting}.
More research is needed to investigate the problem conditions under which such gains occur.

\paragraph{Filtering improves performance.}
Here, we forgo the filtering step, which removes $\approx$ 20\% of the synthesized data on \Yelp.
Table~\ref{tab:ablation} shows that filtering improves the quality of the synthesized data and leads to consistent gains.

\section{Related Work}
\paragraph{Text style transfer} Previous work on style transfer can largely be divided into two categories: (1) learning a latent space with controllable attributes such as those found in \citet{shen2017tyle,john-etal-2019-disentangled} or (2) using unsupervised generative approaches from retrieval \citep{li-etal-2018-delete}, tagging using style phrases \citep{madaan-etal-2020-politeness}, to backtranslation and unsupervised machine translation techniques \citep{prabhumoye2018style,subramanian2019multipleattribute,he2020a}. 

\paragraph{Editing for style transfer}
Our work is closest to \citet{madaan-etal-2020-politeness} and \citet{malmi-etal-2020-unsupervised}.
\citet{madaan-etal-2020-politeness} use a tagger to mark style phrases in the source text, then generates the target text conditioned on the tagged source text.
In contrast, we do not fully generate target text and only perform small, precise edits.
In concurrent work to ours, \citet{malmi-etal-2020-unsupervised} train a BERT language model on each style and edits a span where the models' likelihoods disagree the most.
In contrast, instead of performing single-span replacement, our editor concurrently edits multiple spans in the text, and supports a wider set of operations than replacement.
We showed that this results in more effective and more diverse style transfer.
This coarse-to-fine transformations of text, in which the input context is progressively refined, has also led to improvements in syntactic parsing~\citep{charniak2005coarse}, semantic parsing~\citep{dong2018oarsetofine}, and NER~\citep{choi2018ultra}.

\paragraph{Unsupervised data synthesis for style transfer}
\citet{malmi-etal-2020-unsupervised} also generate synthetic data with which to train an editing model from \citet{malmi-etal-2019-encode}.
Our synthesis differs from~\citet{malmi-etal-2020-unsupervised} in how slots for generation are chosen.
In their work, the highest disagreeing span is chosen for rewriting.
In our work, multiple spans with words whose attention weights that exceed the average are chosen for rewriting, which allows for more flexible and diverse samples.
In turn, training on our synthesized data improves the performance and diversity of the style transfer model.


\section{Conclusion}
We proposed~\mn, a coarse-to-fine editor for style transfer that transforms text using Levenshtein edit operations.
Unlike prior edit methods, our methods concurrently performs multi-span edits.
To train this editor, we proposed an unsupervised data synthesis procedure that converts text to style-agnostic templates using style classifier attention, then fills in slots in these templates using fine-tuned pretrained language models.
\mn~outperformed existing generation and editing style transfer methods on sentiment and politeness transfer.
In addition, the proposed data-synthesis procedure increased transfer performance.
Given the same synthesized data, our editor outperformed prior pure generation and editing methods.
In future work, we will study the application of~\mn~to general sequence to sequence problems.

\section*{Ethical Considerations}
This work has impact in the field of controlled text generation, and as with much of language technology has the potential to be both used for good and used maliciously. Our work learns to generate synthetic data in an unsupervised way, and is based on a pre-trained model (BART), which is likely to caputre and amplity biases found in the data. As with all text-style transfer models, our model is amenable to malicious use, including impersonation and mass generation of faked opposing opinion, for example, negative and positive product reviews or political statements.
\section*{Acknowledgements}
We thank Gabriel Ilharco, Edison Marrese-Taylor, Julian Michaels, Tongshuang Wu, Yutaro Yamada, and Luke Zettlemoyer, and the anonymous reviewers for their helpful feedback on and proofreading of this work. This work was partly supported by the Masason Foundation Fellowship awarded to Machel Reid.
\bibliographystyle{acl_natbib}
\bibliography{acl2021}

\begin{thebibliography}{33}
\expandafter\ifx\csname natexlab\endcsname\relax\def\natexlab#1{#1}\fi

\bibitem[{Charniak and Johnson(2005)}]{charniak2005coarse}
Eugene Charniak and Mark Johnson. 2005.
\newblock \href {https://doi.org/10.3115/1219840.1219862} {Coarse-to-fine
  n-best parsing and {M}ax{E}nt discriminative reranking}.
\newblock In \emph{Proceedings of the 43rd Annual Meeting of the Association
  for Computational Linguistics ({ACL}{'}05)}, pages 173--180, Ann Arbor,
  Michigan. Association for Computational Linguistics.

\bibitem[{Choi et~al.(2018)Choi, Levy, Choi, and Zettlemoyer}]{choi2018ultra}
Eunsol Choi, Omer Levy, Yejin Choi, and Luke Zettlemoyer. 2018.
\newblock \href {https://doi.org/10.18653/v1/P18-1009} {Ultra-fine entity
  typing}.
\newblock In \emph{Proceedings of the 56th Annual Meeting of the Association
  for Computational Linguistics (Volume 1: Long Papers)}, pages 87--96,
  Melbourne, Australia. Association for Computational Linguistics.

\bibitem[{Devlin et~al.(2019)Devlin, Chang, Lee, and
  Toutanova}]{devlin-etal-2019-bert}
Jacob Devlin, Ming-Wei Chang, Kenton Lee, and Kristina Toutanova. 2019.
\newblock \href {https://doi.org/10.18653/v1/N19-1423} {{BERT}: Pre-training of
  deep bidirectional transformers for language understanding}.
\newblock In \emph{Proceedings of the 2019 Conference of the North {A}merican
  Chapter of the Association for Computational Linguistics: Human Language
  Technologies, Volume 1 (Long and Short Papers)}, pages 4171--4186,
  Minneapolis, Minnesota. Association for Computational Linguistics.

\bibitem[{Dong and Lapata(2018)}]{dong2018oarsetofine}
Li~Dong and Mirella Lapata. 2018.
\newblock \href {https://doi.org/10.18653/v1/P18-1068} {Coarse-to-fine decoding
  for neural semantic parsing}.
\newblock In \emph{Proceedings of the 56th Annual Meeting of the Association
  for Computational Linguistics (Volume 1: Long Papers)}, pages 731--742,
  Melbourne, Australia. Association for Computational Linguistics.

\bibitem[{Edunov et~al.(2018)Edunov, Ott, Auli, and
  Grangier}]{edunov-etal-2018-understanding}
Sergey Edunov, Myle Ott, Michael Auli, and David Grangier. 2018.
\newblock \href {https://doi.org/10.18653/v1/D18-1045} {Understanding
  back-translation at scale}.
\newblock In \emph{Proceedings of the 2018 Conference on Empirical Methods in
  Natural Language Processing}, pages 489--500, Brussels, Belgium. Association
  for Computational Linguistics.

\bibitem[{Fu et~al.(2018)Fu, Tan, Peng, Zhao, and Yan}]{fu2017tyle}
Zhenxin Fu, Xiaoye Tan, Nanyun Peng, Dongyan Zhao, and Rui Yan. 2018.
\newblock \href
  {https://www.aaai.org/ocs/index.php/AAAI/AAAI18/paper/view/17015} {Style
  transfer in text: Exploration and evaluation}.
\newblock In \emph{Proceedings of the Thirty-Second {AAAI} Conference on
  Artificial Intelligence, (AAAI-18), the 30th innovative Applications of
  Artificial Intelligence (IAAI-18), and the 8th {AAAI} Symposium on
  Educational Advances in Artificial Intelligence (EAAI-18), New Orleans,
  Louisiana, USA, February 2-7, 2018}, pages 663--670. {AAAI} Press.

\bibitem[{He et~al.(2020{\natexlab{a}})He, Gu, Shen, and
  Ranzato}]{he2020revisiting}
Junxian He, Jiatao Gu, Jiajun Shen, and Marc'Aurelio Ranzato.
  2020{\natexlab{a}}.
\newblock \href {https://openreview.net/forum?id=SJgdnAVKDH} {Revisiting
  self-training for neural sequence generation}.
\newblock In \emph{8th International Conference on Learning Representations,
  {ICLR} 2020, Addis Ababa, Ethiopia, April 26-30, 2020}. OpenReview.net.

\bibitem[{He et~al.(2020{\natexlab{b}})He, Wang, Neubig, and
  Berg{-}Kirkpatrick}]{he2020a}
Junxian He, Xinyi Wang, Graham Neubig, and Taylor Berg{-}Kirkpatrick.
  2020{\natexlab{b}}.
\newblock \href {https://openreview.net/forum?id=HJlA0C4tPS} {A probabilistic
  formulation of unsupervised text style transfer}.
\newblock In \emph{8th International Conference on Learning Representations,
  {ICLR} 2020, Addis Ababa, Ethiopia, April 26-30, 2020}. OpenReview.net.

\bibitem[{He and McAuley(2016)}]{he2016ps}
Ruining He and Julian~J. McAuley. 2016.
\newblock \href {https://doi.org/10.1145/2872427.2883037} {Ups and downs:
  Modeling the visual evolution of fashion trends with one-class collaborative
  filtering}.
\newblock In \emph{Proceedings of the 25th International Conference on World
  Wide Web, {WWW} 2016, Montreal, Canada, April 11 - 15, 2016}, pages 507--517.
  {ACM}.

\bibitem[{Hoover et~al.(2020)Hoover, Strobelt, and
  Gehrmann}]{hoover-etal-2020-exbert}
Benjamin Hoover, Hendrik Strobelt, and Sebastian Gehrmann. 2020.
\newblock \href {https://doi.org/10.18653/v1/2020.acl-demos.22} {ex{BERT}: {A}
  {V}isual {A}nalysis {T}ool to {E}xplore {L}earned {R}epresentations in
  {T}ransformer {M}odels}.
\newblock In \emph{Proceedings of the 58th Annual Meeting of the Association
  for Computational Linguistics: System Demonstrations}, pages 187--196,
  Online. Association for Computational Linguistics.

\bibitem[{John et~al.(2019)John, Mou, Bahuleyan, and
  Vechtomova}]{john-etal-2019-disentangled}
Vineet John, Lili Mou, Hareesh Bahuleyan, and Olga Vechtomova. 2019.
\newblock \href {https://doi.org/10.18653/v1/P19-1041} {Disentangled
  representation learning for non-parallel text style transfer}.
\newblock In \emph{Proceedings of the 57th Annual Meeting of the Association
  for Computational Linguistics}, pages 424--434, Florence, Italy. Association
  for Computational Linguistics.

\bibitem[{Koehn(2004)}]{koehn-2004-statistical}
Philipp Koehn. 2004.
\newblock \href {https://www.aclweb.org/anthology/W04-3250} {Statistical
  significance tests for machine translation evaluation}.
\newblock In \emph{Proceedings of the 2004 Conference on Empirical Methods in
  Natural Language Processing}, pages 388--395, Barcelona, Spain. Association
  for Computational Linguistics.

\bibitem[{Krishna et~al.(2020)Krishna, Wieting, and
  Iyyer}]{krishna-etal-2020-reformulating}
Kalpesh Krishna, John Wieting, and Mohit Iyyer. 2020.
\newblock \href {https://doi.org/10.18653/v1/2020.emnlp-main.55} {Reformulating
  unsupervised style transfer as paraphrase generation}.
\newblock In \emph{Proceedings of the 2020 Conference on Empirical Methods in
  Natural Language Processing (EMNLP)}, pages 737--762, Online. Association for
  Computational Linguistics.

\bibitem[{Lample et~al.(2019)Lample, Subramanian, Smith, Denoyer, Ranzato, and
  Boureau}]{subramanian2019multipleattribute}
Guillaume Lample, Sandeep Subramanian, Eric Smith, Ludovic Denoyer,
  Marc'Aurelio Ranzato, and Y-Lan Boureau. 2019.
\newblock \href {https://openreview.net/forum?id=H1g2NhC5KQ}
  {Multiple-attribute text rewriting}.
\newblock In \emph{International Conference on Learning Representations}.

\bibitem[{{Levenshtein}(1966)}]{Levenshtein}
V.~I. {Levenshtein}. 1966.
\newblock {Binary Codes Capable of Correcting Deletions, Insertions and
  Reversals}.
\newblock \emph{Soviet Physics Doklady}, 10:707.

\bibitem[{Lewis et~al.(2020)Lewis, Liu, Goyal, Ghazvininejad, Mohamed, Levy,
  Stoyanov, and Zettlemoyer}]{lewis-etal-2020-bart}
Mike Lewis, Yinhan Liu, Naman Goyal, Marjan Ghazvininejad, Abdelrahman Mohamed,
  Omer Levy, Veselin Stoyanov, and Luke Zettlemoyer. 2020.
\newblock \href {https://doi.org/10.18653/v1/2020.acl-main.703} {{BART}:
  Denoising sequence-to-sequence pre-training for natural language generation,
  translation, and comprehension}.
\newblock In \emph{Proceedings of the 58th Annual Meeting of the Association
  for Computational Linguistics}, pages 7871--7880, Online. Association for
  Computational Linguistics.

\bibitem[{Li et~al.(2018)Li, Jia, He, and Liang}]{li-etal-2018-delete}
Juncen Li, Robin Jia, He~He, and Percy Liang. 2018.
\newblock \href {https://doi.org/10.18653/v1/N18-1169} {Delete, retrieve,
  generate: a simple approach to sentiment and style transfer}.
\newblock In \emph{Proceedings of the 2018 Conference of the North {A}merican
  Chapter of the Association for Computational Linguistics: Human Language
  Technologies, Volume 1 (Long Papers)}, pages 1865--1874, New Orleans,
  Louisiana. Association for Computational Linguistics.

\bibitem[{Liu et~al.(2019)Liu, Ott, Goyal, Du, Joshi, Chen, Levy, Lewis,
  Zettlemoyer, and Stoyanov}]{liu2019oberta}
Yinhan Liu, Myle Ott, Naman Goyal, Jingfei Du, Mandar Joshi, Danqi Chen, Omer
  Levy, Mike Lewis, Luke Zettlemoyer, and Veselin Stoyanov. 2019.
\newblock Roberta: A robustly optimized bert pretraining approach.
\newblock \emph{arXiv preprint arXiv:1907.11692}.

\bibitem[{Madaan et~al.(2020)Madaan, Setlur, Parekh, Poczos, Neubig, Yang,
  Salakhutdinov, Black, and Prabhumoye}]{madaan-etal-2020-politeness}
Aman Madaan, Amrith Setlur, Tanmay Parekh, Barnabas Poczos, Graham Neubig,
  Yiming Yang, Ruslan Salakhutdinov, Alan~W Black, and Shrimai Prabhumoye.
  2020.
\newblock \href {https://doi.org/10.18653/v1/2020.acl-main.169} {Politeness
  transfer: A tag and generate approach}.
\newblock In \emph{Proceedings of the 58th Annual Meeting of the Association
  for Computational Linguistics}, pages 1869--1881, Online. Association for
  Computational Linguistics.

\bibitem[{Malmi et~al.(2019)Malmi, Krause, Rothe, Mirylenka, and
  Severyn}]{malmi-etal-2019-encode}
Eric Malmi, Sebastian Krause, Sascha Rothe, Daniil Mirylenka, and Aliaksei
  Severyn. 2019.
\newblock \href {https://doi.org/10.18653/v1/D19-1510} {Encode, tag, realize:
  High-precision text editing}.
\newblock In \emph{Proceedings of the 2019 Conference on Empirical Methods in
  Natural Language Processing and the 9th International Joint Conference on
  Natural Language Processing (EMNLP-IJCNLP)}, pages 5054--5065, Hong Kong,
  China. Association for Computational Linguistics.

\bibitem[{Malmi et~al.(2020)Malmi, Severyn, and
  Rothe}]{malmi-etal-2020-unsupervised}
Eric Malmi, Aliaksei Severyn, and Sascha Rothe. 2020.
\newblock \href {https://doi.org/10.18653/v1/2020.emnlp-main.699} {Unsupervised
  text style transfer with padded masked language models}.
\newblock In \emph{Proceedings of the 2020 Conference on Empirical Methods in
  Natural Language Processing (EMNLP)}, pages 8671--8680, Online. Association
  for Computational Linguistics.

\bibitem[{Micikevicius et~al.(2018)Micikevicius, Narang, Alben, Diamos, Elsen,
  Garc{\'{\i}}a, Ginsburg, Houston, Kuchaiev, Venkatesh, and
  Wu}]{micikevicius2018mixed}
Paulius Micikevicius, Sharan Narang, Jonah Alben, Gregory~F. Diamos, Erich
  Elsen, David Garc{\'{\i}}a, Boris Ginsburg, Michael Houston, Oleksii
  Kuchaiev, Ganesh Venkatesh, and Hao Wu. 2018.
\newblock \href {https://openreview.net/forum?id=r1gs9JgRZ} {Mixed precision
  training}.
\newblock In \emph{6th International Conference on Learning Representations,
  {ICLR} 2018, Vancouver, BC, Canada, April 30 - May 3, 2018, Conference Track
  Proceedings}. OpenReview.net.

\bibitem[{Ott et~al.(2019)Ott, Edunov, Baevski, Fan, Gross, Ng, Grangier, and
  Auli}]{ott-etal-2019-fairseq}
Myle Ott, Sergey Edunov, Alexei Baevski, Angela Fan, Sam Gross, Nathan Ng,
  David Grangier, and Michael Auli. 2019.
\newblock \href {https://doi.org/10.18653/v1/N19-4009} {fairseq: A fast,
  extensible toolkit for sequence modeling}.
\newblock In \emph{Proceedings of the 2019 Conference of the North {A}merican
  Chapter of the Association for Computational Linguistics (Demonstrations)},
  pages 48--53, Minneapolis, Minnesota. Association for Computational
  Linguistics.

\bibitem[{Papineni et~al.(2002)Papineni, Roukos, Ward, and
  Zhu}]{papineni-etal-2002-bleu}
Kishore Papineni, Salim Roukos, Todd Ward, and Wei-Jing Zhu. 2002.
\newblock \href {https://doi.org/10.3115/1073083.1073135} {{B}leu: a method for
  automatic evaluation of machine translation}.
\newblock In \emph{Proceedings of the 40th Annual Meeting of the Association
  for Computational Linguistics}, pages 311--318, Philadelphia, Pennsylvania,
  USA. Association for Computational Linguistics.

\bibitem[{Paszke et~al.(2019)Paszke, Gross, Massa, Lerer, Bradbury, Chanan,
  Killeen, Lin, Gimelshein, Antiga, Desmaison, K{\"{o}}pf, Yang, DeVito,
  Raison, Tejani, Chilamkurthy, Steiner, Fang, Bai, and Chintala}]{pytorch}
Adam Paszke, Sam Gross, Francisco Massa, Adam Lerer, James Bradbury, Gregory
  Chanan, Trevor Killeen, Zeming Lin, Natalia Gimelshein, Luca Antiga, Alban
  Desmaison, Andreas K{\"{o}}pf, Edward Yang, Zachary DeVito, Martin Raison,
  Alykhan Tejani, Sasank Chilamkurthy, Benoit Steiner, Lu~Fang, Junjie Bai, and
  Soumith Chintala. 2019.
\newblock \href
  {https://proceedings.neurips.cc/paper/2019/hash/bdbca288fee7f92f2bfa9f7012727740-Abstract.html}
  {Pytorch: An imperative style, high-performance deep learning library}.
\newblock In \emph{Advances in Neural Information Processing Systems 32: Annual
  Conference on Neural Information Processing Systems 2019, NeurIPS 2019,
  December 8-14, 2019, Vancouver, BC, Canada}, pages 8024--8035.

\bibitem[{Post(2018)}]{post-2018-call}
Matt Post. 2018.
\newblock \href {https://doi.org/10.18653/v1/W18-6319} {A call for clarity in
  reporting {BLEU} scores}.
\newblock In \emph{Proceedings of the Third Conference on Machine Translation:
  Research Papers}, pages 186--191, Brussels, Belgium. Association for
  Computational Linguistics.

\bibitem[{Prabhumoye et~al.(2018)Prabhumoye, Tsvetkov, Salakhutdinov, and
  Black}]{prabhumoye2018style}
Shrimai Prabhumoye, Yulia Tsvetkov, Ruslan Salakhutdinov, and Alan~W Black.
  2018.
\newblock \href {https://doi.org/10.18653/v1/P18-1080} {Style transfer through
  back-translation}.
\newblock In \emph{Proceedings of the 56th Annual Meeting of the Association
  for Computational Linguistics (Volume 1: Long Papers)}, pages 866--876,
  Melbourne, Australia. Association for Computational Linguistics.

\bibitem[{Sennrich et~al.(2016)Sennrich, Haddow, and
  Birch}]{sennrich-etal-2016-improving}
Rico Sennrich, Barry Haddow, and Alexandra Birch. 2016.
\newblock \href {https://doi.org/10.18653/v1/P16-1009} {Improving neural
  machine translation models with monolingual data}.
\newblock In \emph{Proceedings of the 54th Annual Meeting of the Association
  for Computational Linguistics (Volume 1: Long Papers)}, pages 86--96, Berlin,
  Germany. Association for Computational Linguistics.

\bibitem[{Shen et~al.(2017)Shen, Lei, Barzilay, and Jaakkola}]{shen2017tyle}
Tianxiao Shen, Tao Lei, Regina Barzilay, and Tommi~S. Jaakkola. 2017.
\newblock \href
  {https://proceedings.neurips.cc/paper/2017/hash/2d2c8394e31101a261abf1784302bf75-Abstract.html}
  {Style transfer from non-parallel text by cross-alignment}.
\newblock In \emph{Advances in Neural Information Processing Systems 30: Annual
  Conference on Neural Information Processing Systems 2017, December 4-9, 2017,
  Long Beach, CA, {USA}}, pages 6830--6841.

\bibitem[{Vaswani et~al.(2017)Vaswani, Shazeer, Parmar, Uszkoreit, Jones,
  Gomez, Kaiser, and Polosukhin}]{vaswani2017ttention}
Ashish Vaswani, Noam Shazeer, Niki Parmar, Jakob Uszkoreit, Llion Jones,
  Aidan~N. Gomez, Lukasz Kaiser, and Illia Polosukhin. 2017.
\newblock \href
  {https://proceedings.neurips.cc/paper/2017/hash/3f5ee243547dee91fbd053c1c4a845aa-Abstract.html}
  {Attention is all you need}.
\newblock In \emph{Advances in Neural Information Processing Systems 30: Annual
  Conference on Neural Information Processing Systems 2017, December 4-9, 2017,
  Long Beach, CA, {USA}}, pages 5998--6008.

\bibitem[{Wolf et~al.(2020)Wolf, Debut, Sanh, Chaumond, Delangue, Moi, Cistac,
  Rault, Louf, Funtowicz, Davison, Shleifer, von Platen, Ma, Jernite, Plu, Xu,
  Le~Scao, Gugger, Drame, Lhoest, and Rush}]{wolf-etal-2020-transformers}
Thomas Wolf, Lysandre Debut, Victor Sanh, Julien Chaumond, Clement Delangue,
  Anthony Moi, Pierric Cistac, Tim Rault, Remi Louf, Morgan Funtowicz, Joe
  Davison, Sam Shleifer, Patrick von Platen, Clara Ma, Yacine Jernite, Julien
  Plu, Canwen Xu, Teven Le~Scao, Sylvain Gugger, Mariama Drame, Quentin Lhoest,
  and Alexander Rush. 2020.
\newblock \href {https://doi.org/10.18653/v1/2020.emnlp-demos.6} {Transformers:
  State-of-the-art natural language processing}.
\newblock In \emph{Proceedings of the 2020 Conference on Empirical Methods in
  Natural Language Processing: System Demonstrations}, pages 38--45, Online.
  Association for Computational Linguistics.

\bibitem[{Wu et~al.(2019)Wu, Zhang, Zang, Han, and Hu}]{wu2019mask}
Xing Wu, Tao Zhang, Liangjun Zang, Jizhong Han, and Songlin Hu. 2019.
\newblock \href {https://doi.org/10.24963/ijcai.2019/732} {Mask and infill:
  Applying masked language model for sentiment transfer}.
\newblock In \emph{Proceedings of the Twenty-Eighth International Joint
  Conference on Artificial Intelligence, {IJCAI} 2019, Macao, China, August
  10-16, 2019}, pages 5271--5277. ijcai.org.

\bibitem[{Zhang et~al.(2020)Zhang, Kishore, Wu, Weinberger, and
  Artzi}]{zhang2020bert}
Tianyi Zhang, Varsha Kishore, Felix Wu, Kilian~Q. Weinberger, and Yoav Artzi.
  2020.
\newblock \href {https://openreview.net/forum?id=SkeHuCVFDr} {Bertscore:
  Evaluating text generation with {BERT}}.
\newblock In \emph{8th International Conference on Learning Representations,
  {ICLR} 2020, Addis Ababa, Ethiopia, April 26-30, 2020}. OpenReview.net.

\end{thebibliography}
\appendix
\section{Evaluation}
\paragraph{Automatic Evaluation} When evaluating using BLEU we used detokenized SacreBLEU \citep{post-2018-call}. For BERTScore \citep{zhang2020bert}, we use the \texttt{rescale\_with\_baseline} option with the following hash: \texttt{roberta\-large\_L17\_no-idf\_version=0.3.5(hug\_\\trans=4.1.0.dev0)-rescaled}.
\paragraph{Human Evaluation} When using Amazon Mechanical Turk, we screen our annotators for English proficiency, and require all to have a greater than 95\% approval rate. We hire workers with at least 1000 HITs and pay workers 5 cents per example, amounting to USD\$10-15 per hour.
\section{Training Infrastructure}
For training models we use between 1 and 8 NVIDIA V100 16GB GPUs on a DGX-1 machine running Ubuntu 16.04 on a Dual 20-Core Intel Xeon E5-2698 v4 2.2 GHz.
\section{Source Code \& Synthetic Data}
We release source code with this work, with preprocessing scripts, training scripts for both  conditional lanaguage models, editors and coarse-grain taggers, edit operations extraction scripts, and synthetic data generation scripts at \url{https://github.com/machelreid/lewis}.

\section{Synthetic Data}
For synthetic data generation, we generate approximately 2.2M pairs for Yelp, 2.0M pairs for Amazon, and 1M pairs for Polite. Note that when generating synthetic data on Polite, given the longer sequence length, we threshold the amount of \texttt{SLOT} tokens to be the minimum of one-third of the total sequence length and 6.
\\~\\We release our synthetic data to help facilitate further development in approaches using synthetic data for this task.

\section{Qualitative Analysis}
We analyzed 100 examples from YELP produced by \mn. 83\% transfers were correct,6\% incorrect,and 11\% ambiguous (the resulting sentence expressed both styles). This is in line with automatic metrics and shows \mn~is effective in successfully transferring style. For diversity of edits, in 59\% of cases, \mn~inverted key phrases (\textit{and enjoying this} $\rightarrow$ \textit{and avoiding this}, \textit{friendly folks, delicious authentic bagels} $\rightarrow$ \textit{sorry folks,not authentic bagels}), in 26\%, \mn~rewrote part of the sentence in a way that is not inverting key adjectives/nouns (and he loved it $\rightarrow$ and he said it was OK). In 10\%, \mn~performed purely syntactic editing (\textit{definitely not enough room} $\rightarrow$ \textit{enough room}). In contrast to other editors that rely on primarily single-phrase inversion (e.g. LaserTagger), demonstrating that \mn~provides diverse edits.

\section{Further Automatic Evaluation}
We further evaluate our model on semantic similarity and fluency using the classifiers released by \citet{krishna-etal-2020-reformulating}. Results are shown in Table~\ref{tab:fluency_class} and \ref{tab:sim_class}. \mn~improves fluency by a significant margin on all, and outperforms other methods on 2/3 datasets on semantic similarity.
\begin{table}[h]
    \centering
    \footnotesize
    \resizebox{!}{!}{
    \begin{tabular}{lcccc}
    \toprule
         \textbf{Dataset} &\textbf{Model} & \textbf{Fluency} \\
         \midrule
         \multirow{5}{*}{\Yelp} 
         & Delete and Retrieve & 38.7 \\
         & TG & 53.1 \\
         & DeepLatentSeq & 68.1 \\
         & \mn & 84.5\\
         & Gold & 89.4 \\
         \midrule
         \multirow{4}{*}{\Amazon}
         & Delete and Retrieve & 49.2 \\
         & TG & 54.6 \\
         & \mn & 85.6 \\
         & Gold & 84.5 \\
         \midrule
         \multirow{2}{*}{\Polite} 
         & TG & 67.5 \\
         & \mn~& 93.0  \\
         & Gold (src) & 92.3  \\ 
         \bottomrule
    \end{tabular}}
    \caption{Fluency classification from \citet{krishna-etal-2020-reformulating}}
    \label{tab:fluency_class}
\end{table}
\begin{table}[t]
    \centering
    \footnotesize
    \resizebox{0.5\textwidth}{!}{
    \begin{tabular}{lcccc}
    \toprule
         \textbf{Dataset} &\textbf{Model} & \textbf{Target sim.} & \textbf{Source sim.} \\
         \midrule
         \multirow{5}{*}{\Yelp} 
         & Delete and Retrieve& 49.2& 61.0 \\
         & Deep Latent Seq& 48.8& 60.5 \\
         & TG& 54.0& 67.3 \\
         & \mn& 61.2& 77.4 \\
         & Gold & --- &65.7 \\
         \midrule
         \multirow{4}{*}{\Amazon} 
          & Delete and Retrieve& 64.9& 81.0 \\
          & TG& 59.2 & 76.3 \\
          & \mn& 59.6& 78.8 \\
          & Gold & --- & 75.7 \\ 
         \midrule
         \multirow{2}{*}{\Polite} 
         & TG & --- & 84.6 \\
         & \mn & --- & 87.4\\
         \bottomrule
    \end{tabular}}
    \caption{Semantic similarity classification from \citet{krishna-etal-2020-reformulating}}
    \label{tab:sim_class}
\end{table}
\section{Example Outputs}
\begin{table}[H]
    \centering
    \scalebox{0.9}{\begin{tabularx}{\linewidth}{lX}
    \toprule[0.1em]
         \textbf{Source} & the wine was very average and the food was even less .\\
         \textbf{\mn} & the wine was very good and the food was even less . \\
         \textbf{LaserTagger} & the wine was very good and the food was even better .	\\
          \textbf{Reference} &  the wine was above average and the food was even better\\
          \midrule
          \textbf{Source} & owner : a very rude man .\\
          \textbf{\mn} &    owner : a very nice man .\\
          \textbf{LaserTagger} & owner : a very man .\\
          \textbf{Reference} & The owner was such a friendly person.\\
          \midrule
          \textbf{Source} &i love the food ... however service here is horrible .\\
          \textbf{\mn} &  i  love the food and the service here is great . \\
          \textbf{LaserTagger} & i love the food ... however service here is great .\\
          \textbf{Reference} & i love the food ... and service here is awesome .\\

        \bottomrule[0.1em]
    \end{tabularx}}
    \caption{Three examples from the Yelp test set comparing the LaserTagger trained on our synthetic data and \mn}
    \label{app:comparison}
\end{table}
\end{document}